# ANALYSING QUALITY OF ENGLISH-HINDI MACHINE TRANSLATION ENGINE OUTPUTS USING BAYESIAN CLASSIFICATION

Rashmi Gupta[1], Nisheeth Joshi[2] and Iti Mathur[3]

[1,2,3] Apaji Institute, Banasthali University, Rajasthan, India,

[1] `rsh.gupta06@gmail.com`, [2] `nisheeth.joshi@rediffmail.com`,

[3] `mathur_iti@rediffmail.com`

## ABSTRACT

*This paper considers the problem for estimating the quality of machine translation outputs which are independent of human intervention and are generally addressed using machine learning techniques. There are various measures through which a machine learns translations quality. Automatic Evaluation metrics produce good co-relation at corpus level but cannot produce the same results at the same segment or sentence level. In this paper 16 features are extracted from the input sentences and their translations and a quality score is obtained based on Bayesian inference produced from training data.*

## 1. INTRODUCTION

Quality Estimation(Confidence Estimation) can be seen as a binary classification problem for judging the quality of the output sentences. It correlates better with human evaluations. The task of machine translation evaluation can be distinguish from the confidence estimation. The goal of machine translation system is to compare the machine translation to a reference translation and to check how close is the machine translation to a reference translations while in the case of confidence estimation the task is to predict the quality of the translated sentences without any information about the expected output. In this paper a Naïve Bayes classifier is trained to predict different types of sentence level scores. The remaining section of the paper is organised as: Section 2 gives the previous work on confidence estimation for machine translation. Section 3 describes the system description. Section 4 gives the experimental settings. Section 5 gives the comparision of human evaluation and Naïve bayes. Section 6 finally gives the conclusion.

            

## 2. RELATED WORK

Most of the work on the sentence level Quality estimation focused on estimating the general quality score and also on estimating the post editing effort. Cortson-Oliver et.al. [1] trained a classifier to distinguish between machine translations and human translations using language models. Kulesza and shieber [2] also did the same work by using support vector machines classifiers and also uses the features which are based on machine translation evaluation metrics such as WER, PER, BLEU and NIST. Blatz et.al. [3] did the first sentence level quality estimation of machine translation using 91 features. Gamon et al. [4] trained an SVM classifier using a number of linguistic features which were extracted from machine and human translations to differenciate between human and machine translations. Albrecht and Hwa [5] used a regression algorithm with string-based and syntax-based features which were extracted from Machine Translation output. Pado et.al. [6] used a regression algorithm along with features which contained textual entailment between the translation and the reference sentences. Specia.et.al. [7] used 74 features to trained a support vector machine classifier. Specia.et.al [8] used the more linguistic features like POS tags, chunks, dependency parses and name entities on English-Arabic Quality estimation for machine translations.

| Code | MT System |
|------|-----------|
| E1 | Google Translator |
| E2 | Bing Translator |
| E3 | Babylon Translator |

Table 1. English-Hindi Machine Translatiors used in the study

## 3. SYSTEM DESCRIPTION

In this section the details of our system is specified. We have worked on 16 features. The features are first extracted from the source sentences and their corresponding translations from the corpus. The goal of supervised learning is to predict the class labels of examples that have not been seen. The trained model is first trained on the corpus with labeled quality score and then it is able to predict the score for unlabeled sentences.

### 3.1. Naïve Bayes

Naïve bayes is a well known algorithm for classification problems. The list of the sets of attribute values and its corresponding category are given to the classifier and these constitue the training set. From the training data an independent probability is established. The probability gives the likelihood of each target class, given the occurrence of each value category from each input variable. When a new example is presented a value for the target function can be predicted based on the training instances. A joint distribution over a label Y and a set of observed features $(f_1, f_2, f_n)$ using assumption that the full joint distribution can be factored as follows:

$$P(f1, f2 \ldots fn, y) = P(y) \prod_{i=1}^{n} P(Fi|y) \tag{1}$$

P(C) is estimated by the relative frequencies of the training pairwise examples. The naïve bayes has the advantage that it is computationally fast and scalable that calculates conditional probabilities for combination of attributes and target attributes.

## 3. EXPERIMENTAL SETUP

For development of the training system, we used a 3,300 sentence corpus that was built during ACL 2005 workshop on building and using parallel text : Data Driven machine translation and beyond,as the training corpus. The statistics of the corpus is shown in table2.

| Corpus | English-Hindi Parallel Corpus |
|---|---|
| Sentences | 3,300 |
| Words | 55,014 |
| Unique Words | 8,956 |

Table 2 Statistics of training corpus

In this paper we have focused on using supervised machine learning in evaluation of MT engine outputs without using human reference translations. WEKA toolkit is used for training this classifier. We have trained a Naïve Bayes classifier. 16 features were used for training our classifiers. These features were as follows:

| S. No. | Features Description |
|---|---|
| 1 | Number of tokens in the source sentence. |
| 2 | Number of tokens in the Target sentence. |
| 3 | Average source token length. |
| 4 | Language model probability of source sentence. |
| 5 | Language model probability of Target sentence. |
| 6 | Average number of occurrence of target words within the target sentence. |
| 7 | Average number of translation per source word in the sentence. |
| 8 | Percentage of low frequency unigram in the source language. |
| 9 | Percentage of high frequency unigram in the source language. |
| 10 | Percentage of low frequency bigrams in the source language. |
| 11 | Percentage of high frequency bigrams in the source language. |
| 12 | Percentage of high frequency trigrams in the source language. |
| 13 | Percentage of low frequency trigrams in the source language |
| 14 | Percentage of unigrams in the source sentence seen in the corpus |
| 15 | Number of punctuation marks in the source sentence |
| 16 | Number of punctuation marks in the target sentence |

Table 3. Feature set used in training the supervised model

The output from the training corpus were registered against all these three machine translation engines and a human evaluator was asked to judge the outputs. The judging criteria was same as used by Joshi et. al. [9]. All the sentences were judged on ten parameters using a scale between 0-4. The ten parameters used in evaluation are shown in table 4.

Once the human evaluation of these outputs were done, we used these results along with the 16 features that were extracted from the English source sentences and Hindi MT outputs. We tested the classifiers using another corpus of 1300 sentences. Table 6 shows the statistics of this corpus. These 1300 sentences were divided into 13 documents of 100 sentences each. We registered the outputs of the test corpus on all three MT engines and performed human evaluation on them.

| S.No. | Parameter |
|---|---|
| 1 | Translation of Gender and Number of the Noun(s). |
| 2 | Identification of the Proper Noun(s). |
| 3 | Use of Adjectives and Adverbs corresponding to the Nouns and Verbs. |
| 4 | Selection of proper words/synonyms (Lexical Choice). |
| 5 | Sequence of phrases and clauses in the translation. |
| 6 | Use of Punctuation Marks in the translation |
| 7 | Translation of tense in the sentence |
| 8 | Translation of Voice in the sentence |
| 9 | Maintaining the semantics of the source sentence in the translation |
| 10 | Fluency of translated text and translator's proficiency |

Table 4. parameters used in evaluation

## 5. Comparison of Human Evaluation and Naïve Bayes

For the evaluation of the system, we converted the human evaluation of the systems into grades. These grades were converted using table 7. Based on these grades we computed the results of the classifier, which gave us the same classes. Table 8 shows the results produced by naive bayes classifier and human evaluation. This give the number of times an MT engine scored in either of the four categories. Table 9 shows the results of human evaluation for the three MT engines. These four grades can also be converted into a numeric score to provide ranks to the MT outputs. Table 10 shows the comparison of results of human grades with the grades given by the classifier. Here, we counted the number of times the same grade was provided by the human evaluators as well as the classifier to a given sentence i.e. if a human evaluator gave a good score to a sentence and the classifier also gave good to the same sentence then will counted it. Thus from these tables we have shown that the Naïve Bayes classifier can predict the same level of outputs as that of human evaluator. More the human evaluator and the classifier can produce almost similar results to most of the judgments.

| Score | Description |
|---|---|
| 1 | Ideal |
| 2 | Perfect |
| 3 | Acceptable |
| 4 | Partially Acceptable |
| 5 | Not Acceptable |

Table 5. Interpretation of Human Evaluation Scale 5

## 6. CONCLUSIONS

The assessment of the quality of the sentences can also be done by a human expert but this is very tedious and time consuming task and may not be possible if the expert does not have the good knowledge of the source sentences. In this paper we have extracted 16 features from the input sentences and their translation and a quality score is obtained based on Bayesian inference produced from training data. Our system gives score and correlates well with human evaluation.

| Corpus | English corpus |
|---|---|
| sentences | 1300 |
| words | 26,724 |
| Unique words | 3,515 |

Table 6. Statistics for Test Corpus

| S.NO | Score Range | Grade |
|---|---|---|
| 1 | 0 – 0.250 | Poor |
| 2 | 0.251 – 0.50 | Average |
| 3 | 0.51 – 0.75 | Good |
| 4 | 0.751 – 1.0 | Excellent |

Table 7 Grade allocated to Human Evaluation

| Grade | Bing | Google | Babylon |
|---|---|---|---|
| Excellent | 24 | 23 | 12 |
| Good | 228 | 221 | 200 |
| Average | 1019 | 1008 | 1025 |
| Poor | 29 | 48 | 63 |

Table 8. Naïve Bayes Classifiers Results

| Grade | Bing | Google | Babylon |
|---|---|---|---|
| Excellent | 96 | 92 | 7 |
| Good | 231 | 194 | 234 |
| Average | 956 | 1002 | 1006 |
| Poor | 17 | 12 | 53 |

Table 9: Human Evaluation Results

| S.NO | MT Engine | Same Results | Percentage |
|---|---|---|---|
| 1 | Bing | 771 | 59.30 |
| 2 | Google | 756 | 58.15 |
| 3 | Babylon | 711 | 54.69 |

Table10: Comparision of Naïve Bayes Classifier and Human Evaluator's Results

**Authors**

Rashmi Gupta is pursuing her M.Tech in Computer Science From Banasthali University, Rajasthan and is working as a Research Assistant in English- Indian Languages Machine Translation System Project sponsored by TDIL Programme, DEITY. She has developed an Automatic Ranking System for Outputs of English -Hindi Machine

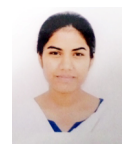

Translation Systems. Her current research interest includesNatural Language Processing and Machine Translation.

Nisheeth Joshi is a researcher working in the area of Machine Translation. He has been working in design and development of evaluation Matrices in Indian Languages. Besides this he is also actively involved in the development of MT engines for English to Indian Languages. He has several publications in various journals and conferences and also serves on the ProgrammeCommittees and Editorial Boards of several conferences and journal 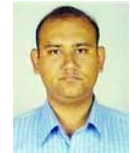

ItiMathur is an assistant professor at Banasthali University. Her primary area of research is computational semantics and ontological engineering. Besides this she is also involved in the development of MT engines for English to Indian Languages. She is one of the experts empanelled with TDI Programme, Department of Electronics and Information Technology(DeitY), Govt.of India, a premier organization which foresees Language Technology Funding and Research in India. She has several publications in various journals and conferences and also serves on the Programme Committees and Editorial Boards of several conferences and journal. 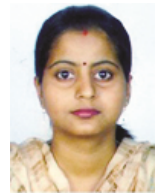